\documentclass[a4paper, 10pt, conference]{ieeeconf}      

\IEEEoverridecommandlockouts                              
                                                          
\usepackage{url}
\usepackage{comment}
\usepackage{amsmath,amssymb,amsfonts, mathtools}
\usepackage{graphicx}
\usepackage{bm}
\usepackage{epstopdf}
\usepackage{placeins}
\usepackage{bm}
\usepackage{cite}
\let\labelindent\relax
\usepackage[inline]{enumitem}
\usepackage{multicol}
\usepackage{tikz}
\usepackage{caption}
\usepackage{subcaption}
\usepackage{tikz}
\usepackage{svg}

\newcommand{\vect}[1]{\boldsymbol{#1}}
\newcommand\numberthis{\addtocounter{equation}{1}\tag{\theequation}}

\title{\LARGE \bf Confidence-Based Skill Reproduction Through Perturbation Analysis}

\author{Brendan Hertel and S. Reza Ahmadzadeh
	\thanks{Persistent Autonomy and Robot Learning (PeARL) Lab, University of Massachusetts Lowell, MA, 01854. Email: \tt{brendan\_hertel@student.uml.edu,reza@cs.uml.edu}}
 }
 
\begin{document}

\maketitle
\thispagestyle{empty}
\pagestyle{empty}

\begin{abstract}
Several methods exist for teaching robots, with one of the most prominent being Learning from Demonstration (LfD). Many LfD representations can be formulated as constrained optimization problems. We propose a novel convex formulation of the LfD problem represented as elastic maps, which models reproductions as a series of connected springs. Relying on the properties of strong duality and perturbation analysis of the constrained optimization problem, we create a confidence metric. Our method allows the demonstrated skill to be reproduced with varying confidence level yielding different levels of smoothness and flexibility. Our confidence-based method provides reproductions of the skill that perform better for a given set of constraints. By analyzing the constraints, our method can also remove unnecessary constraints. We validate our approach using several simulated and real-world experiments using a Jaco2 7DOF manipulator arm.
%
\end{abstract}

\section{Introduction}
\label{sec:intro}

One of the most efficient methods for teaching robots new skills is Learning from Demonstration (LfD) where a teacher shows a skill to a robot in order to enable it to reproduce the skill under new constraints~\cite{Argall2009survey}. Many LfD representations can be formulated as an optimization problem, where some cost function is formulated and minimized to find a reproduction while satisfying any set of given constraints. Some representations explicitly minimize a pre-defined cost function~\cite{hertel2021TLFSD, Ravichandar2019MCCB}. For example, \emph{Trajectory Learning via Failed and Successful Demonstrations} (TLFSD)~\cite{hertel2021TLFSD} minimizes distance from successful demonstrations while maximizing distance from failed ones. Similarly \emph{Multi-Coordinate Cost Balancing}~\cite{Ravichandar2019MCCB} optimizes between several differential coordinate representations of demonstrations to find the optimal reproduction. Other representations are not formulated as optimization problems, but can be interpreted as such. \emph{Laplacian Trajectory Editing}~\cite{Nierhoff2016LTE} transforms the demonstrations into curvature space, applies constraints, and then inverts the transform back into Cartesian space. This algorithm formulates the LfD problem as a least squares optimization. Alternatively, \emph{Dynamic Movement Primitives} (DMPs)~\cite{pastorDMP2009} use a dynamical system to model and execute the reproduction. Solving this dynamical system can also be formulated as an optimization problem involving Hilbert norm minimization~\cite{dragan2015movement}.

\begin{figure}[t]
    \centering
    \includegraphics[width=0.9\linewidth]{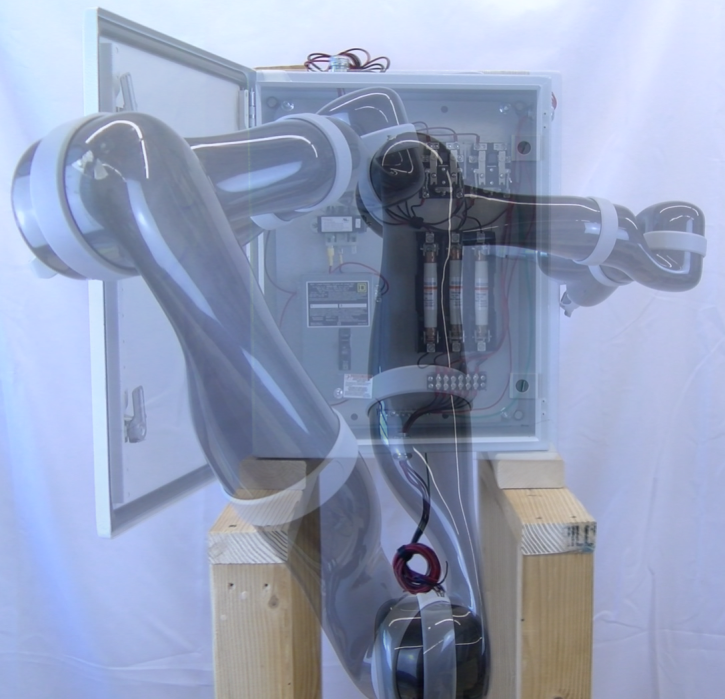}
    \caption{\small Reproduction of a door opening task using a level of confidence which results in success.}
    \label{fig:fig1}
\end{figure}

\begin{figure*}[t]
    \centering
    \includegraphics[width=0.9\linewidth]{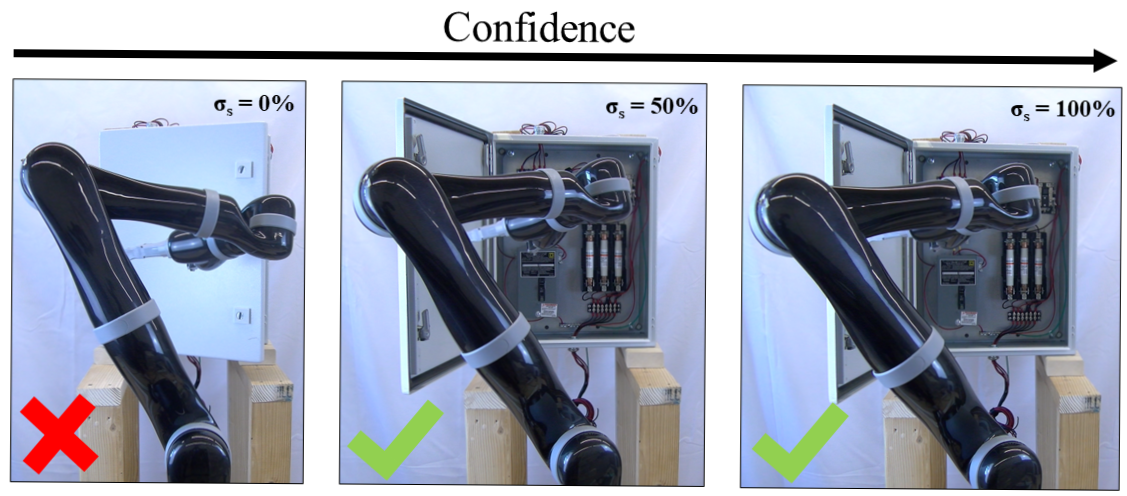}
    \caption{\small Demonstration and reproductions with different confidence factors of opening a real-world box. As confidence in a reproduction increases, the constraints tighten. Left: a low-confidence reproduction does not successfully open the box. Center and right: higher confidence reproductions successfully complete the task with different features.}
    \label{fig:fig2}
\end{figure*}

Optimization problems, particularly convex optimization problems, have certain properties which to our knowledge have not yet been extensively exploited for LfD~\cite{boyd2004convex}. One property of interest is \emph{perturbation analysis} that focuses on how the optimal value changes when constraints are perturbed. In this paper, we propose a new convex formulation of the LfD problem using elastic maps~\cite{hertel2022ElMap}. Then we derive the dual problem to use duality conditions and perturbation analysis to exploit knowledge about the reproductions, creating a confidence metric for each reproduction. This method presents several advantages over previous methods. First is that our method provides confident reproductions of the skill for a specific set of constraints. Confident reproductions, such as the one shown in Fig.~\ref{fig:fig1}, perform better at the intended skill for the given constraints. Additionally, since more confident reproductions have tighter constraints, which does not allow for variability and smoothness in reproductions, the confidence in a reproduction may be tuned as shown in Fig.~\ref{fig:fig2} to allow for reproductions of different smoothness and variability. Unlike other methods that consider static constraints, another advantage of our method is it analyzes constraints and considers perturbations of these constraints. Through constraint analysis, our method can remove unnecessary constraints using duality and properties of the dual values of constraints. We validate our approach in two simulated and one real-world experiment using a Jaco 7DOF manipulator arm.


\section{Related Work} 
\label{sec:RW}

Learning from Demonstration covers a large variety of techniques, which can be in the form of dynamical~\cite{pastorDMP2009}, statistical~\cite{hertel2022ElMap}, geometric~\cite{Nierhoff2016LTE,
ahmadzadeh2018trajectory, ahmadzadeh2017generalized}, or probabilistic methods~\cite{Paraschos2013ProMP, rana2018learning}. Each of these different types of methods present different advantages when reproducing trajectories~\cite{hertel2021SAMLfD}. While some methods rely on explicit optimization~\cite{hertel2021TLFSD, hertel2022ElMap, dragan2015movement}, others can be reformulated as optimization problems~\cite{pastorDMP2009, Nierhoff2016LTE}. Many explicit optimization formulations use a cost function and rely on tradeoffs between elements of the cost function to reproduce demonstrations. When optimizing, using different norms results in different properties of the reproduction, such as different velocity or jerk~\cite{dragan2015movement}. The Jerk-Accuracy method uses parameter is used to tradeoff between converging to a given demonstration and minimizing jerk~\cite{Meirovitch2016JA}. Other methods do not use tradeoff values, instead optimizing without parameters. Rana et al. proposed a method which incorporates both LfD and motion planning techniques~\cite{rana2018clamp}. This method uses a \textit{maximum a posteriori} inference to create trajectories, using effective optimization strategies designed for motion planning~\cite{dong2016motion}. Alternatively, multimodal approaches combine several optimization approaches to create a family of reproductions which are optimal for the skill~\cite{osa2020multimodal}. Other LfD representations may not explicitly use optimization but can be formulated as an optimization problem. As shown in \cite{dragan2015movement}, DMPs~\cite{pastorDMP2009} can be formulated as optimizing trajectory reproduction using a special case of Hilbert norm minimization. While these previous works use optimization either explicitly or implicitly, they do not take advantage of certain properties associated with optimization such as perturbation analysis. Additionally, not all works formulate convex optimization problems, which can be solved efficiently and have strong duality. Non-convex problems must be solved by an optimal search algorithm, and do not provide valuable information regarding constraints.

Various methods have been used to increase confidence in the execution of a robot's movements. In \cite{hertel2021SAMLfD}, several LfD methods are evaluated using a similarity metric, and the highest similarity reproduction is selected. In \cite{cao2021learning}, the authors propose a method to avoid unsuccessful reproductions in which a reproduction achieving the highest ``feasibility score'' for the imitator is performed. Other methods focused on avoiding failure have been utilized, such as ergodic imitation or safety credit assignment. Ergodic imitation~\cite{kalinowska2021ergodic} relies on reproducing demonstrations with maximum ergodicity, resulting in a successful reproduction. Safety credit assignment~\cite{prabhakar2021credit} learns a safety barrier from demonstrations, which can then be used with a controller to prevent unsafe executions. Alternatively, confidence-based autonomy~\cite{Chernovaconfidence2009} promotes more efficient learning from demonstrations by requesting demonstrations when confidence in reproductions is low. This method also incorporates skill refinement, which allows a user to correct portions of the reproduction. Skill refinement allows users to increase the likelihood of success in reproductions by adjusting reproductions until they are suitable for the skill~\cite{ argall2010tactile}. Our perturbation analysis approach is similar to these methods as it uses spatial properties of the demonstrations to measure confidence in the reproduction, but also allows for reproductions of varying confidence levels, as sometimes reproductions of lower confidence may be more desirable. Perturbation analysis also allows for visualizing confidence across the reproduction space.

\section{Background}
\label{sec:bg}

\subsection{Elastic Maps}
Elastic maps are a tool for nonlinear dimensionality reduction. They can be used to approximate a lower-dimensional manifold of given data~\cite{gorban_zinovyev}. Elastic maps have been used for data analysis and visualization across several fields, including bioinformatics~\cite{gorban_zinovyev}, political science, and social science~\cite{gorban_visualization2001}. 
Maps are represented by nodes, which are interconnected by edges, and adjacent edges form a rib. Elastic maps have a mechanical interpretation inspired by springs. Nodes connect to the data and other nodes through springs. By minimizing three energies associated with the elastic map and its spring-like structure, an optimal representation of the data is found. The three energies are
\begin{enumerate*}[label=(\roman*)]
  \item the approximation energy $U_\mathcal{X}$ which penalizes a bad fit to the data,
  \item the stretching energy $U_E$ which penalizes high distance between adjacent nodes, and
  \item the bending energy $U_R$ which penalizes the curvature of nodes.
\end{enumerate*}
These energies may be formulated differently depending upon the structure of the elastic map.

\subsection{Elastic Maps for Trajectory Reproduction}
We have previously developed an approach using elastic maps for the use of trajectory reproduction in \cite{hertel2022ElMap}. Of particular use for trajectories are polyline elastic maps, where each node has only two edges except for two terminal nodes. The energy terms in a polyline elastic map can be defined as follows:
\begin{align}
    U_\mathcal{X} &= \frac{1}{\sum_{\zeta_j} w_j} \sum_{i = 1}^{N} \sum_{\zeta_j \in k_i} w_j || \zeta_j - x_i ||_2^2, \label{eq:U_X}\\
    U_E &= \alpha \sum_{i = 1}^{N-1} || x_{i+1} - x_i ||_2^2,  \label{eq:U_E}\\
    U_R &= \beta \sum_{i = 1}^{N-2} || x_i - 2 x_{i+1} + x_{i+2} ||_2^2, \label{eq:U_R}
\end{align}
where $\vect{\zeta} = [\zeta_1, \zeta_2, ..., \zeta_M]$ is the demonstration data, $w_j$ is the weight of data $\zeta_j$, $\vect{x} = [x_1, x_2, ..., x_N]$ is the nodes in the elastic map, $k_i$ is the cluster of data for node $x_i$, $||\cdot||_n$ is the $L^n$-norm, and $\alpha$ and $\beta$ are the stretching and bending constants, respectively.

Optimizing these three energies results in a series of nodes which has properties particularly well-suited to a robot trajectory; a smooth, evenly-spaced reproduction fitted to the demonstrations which adheres to all given constraints. Performance using elastic maps compared to other contemporary methods is shown in \cite{hertel2022ElMap}. Additionally, elastic maps can be used with any number of demonstrations and/or constraints, which provides flexibility in its applications for robot trajectories. In this paper we utilize elastic maps using convex optimization techniques 
and analyze how perturbations of constraints can affect the optimal solution to understand confidence in elastic map reproductions of robot trajectories. Elastic maps are used in favor of other methods due to their flexibility with the number of demonstrations and constraints. 

\section{Methodology} 
\label{sec:method}

\subsection{Problem Formulation: Derivation of the Primal}

To find an optimal elastic map representation of demonstration data, the following optimization problem is solved:
\begin{align*}
    &\underset{\vect{x}}{\text{minimize }}  f_0(\vect{x}) = U_\mathcal{X} + U_E + U_R \numberthis\label{eq:opt} \\
    &\text{subject to} \ f_i(\vect{x}) = ||y - x_j||_1 - r \leq 0
\end{align*}
where $f_i(\vect{x})$ is the $i$th constraint, taking the form of constraining a point on the reproduction $x_j$ to some point in space $y$ within radius $r$. The form of these constraints is discussed further in Sec.~\ref{subsec:PA}. To use convex minimization efficiently, the energy formulations are changed as follows:
\begin{align}
    U_\mathcal{X} &= \gamma  || \vect{I}\vect{x} -\vect{K}\vect{\zeta} ||_2^2 \label{eq:U_Y_new}\\
    U_E &= \alpha  || \vect{E}\vect{x} ||_2^2  \label{eq:U_E_new}\\
    U_R &= \beta  || \vect{R}\vect{x} ||_2^2, \label{eq:U_R_new}
\end{align}
where $\gamma = (\sum_{\zeta_j} w_j)^{-1}$ is the normalization factor for the data weights, $\vect{I}$ is the identity matrix, $\vect{K}$ is a clustering matrix which incorporates the individual weights of data points, $\vect{E}$ and $\vect{R}$ define the edges and ribs such that
\begin{equation}
\vect{E} = 
\begin{bmatrix}
 -1 & 1 &  0 & \cdots & 0 \\
0 &  -1 & 1 & \cdots & 0 \\
\vdots & \ddots &\ddots & \ddots & \vdots \\
0 & \cdots & 0 & -1 & 1   \\
\end{bmatrix}, \\ \nonumber \\
\end{equation}
\begin{equation}
\vect{R} = 
\begin{bmatrix}
1 & -2 &  1 & 0 & \cdots & 0 \\
0 & 1 & -2 & 1 & \cdots & 0 \\
\vdots & \ddots & \ddots & \ddots & \ddots & \vdots\\
0 & \cdots & 0 & 1 & -2 & 1   \\
\end{bmatrix} \nonumber
\end{equation}
which results in a convex minimization that can be solved efficiently. The use of $\vect{I}$ in the above equations is only permissible when $\vect{x}$ and $\vect{\zeta}$ are the same size and uniform weights are used. In all other cases, a diagonal weighted matrix must be used instead. Note that unlike \cite{hertel2022ElMap}, constraints will not be included in $\vect{\zeta}$. Constraints included in $\vect{\zeta}$ are forced to be met exactly, while convex constraints include inequalities. This allows for further flexibility in reproductions, and makes properties such as obstacle avoidance easier to implement. Without the convex optimization, obstacle avoidance must be used with a specific via-point or set of via-points.
With convex formulation, obstacle avoidance can be implemented automatically without applying hard constraints to the reproduction.

\subsection{Problem Formulation: Derivation of the Dual}
We derive the Lagrangian dual problem for the primal in \eqref{eq:opt}. We first  expand \eqref{eq:opt} as follows:
\begin{align}
    f_0(\vect{x}) &= \gamma  || \vect{I}\vect{x} -\vect{K}\vect{\zeta} ||_2^2 + \alpha  || \vect{E}\vect{x} ||_2^2 + \beta  || \vect{R}\vect{x} ||_2^2 \nonumber \\   
    & =  \gamma  (\vect{x}^\top \vect{I}^\top \vect{I} \vect{x} + (\vect{K} \vect{\zeta})^\top (\vect{K} \vect{\zeta}) - 2 (\vect{K} \vect{\zeta})^\top (\vect{I} \vect{x}) ) \nonumber\\ & + \alpha\vect{x}^\top \vect{E}^\top \vect{E} \vect{x} + \beta\vect{x}^\top \vect{R}^\top \vect{R} \vect{x},
\end{align}
\noindent which can be simplified to
\begin{align}
    f_0(\vect{x}) &=  \vect{x}^\top (\gamma\vect{I}^\top \vect{I} + \alpha\vect{E}^\top \vect{E} + \beta \vect{R}^\top \vect{R}) \vect{x} \nonumber\\&- \gamma(\vect{K} \vect{\zeta})^\top (2 \vect{I} \vect{x} - \vect{K} \vect{\zeta} ). \label{eq:vectorized}
\end{align}
The Lagrangian then can be calculated by combining~\eqref{eq:vectorized} and the inequality constraints in \eqref{eq:opt} as follows:
\begin{align}
     \mathcal{L}(\vect{x}, \vect{\lambda}) = & \vect{x}^\top (\gamma\vect{I}^\top \vect{I} + \alpha\vect{E}^\top \vect{E} + \beta \vect{R}^\top \vect{R}) \vect{x} \nonumber\\& - \gamma(\vect{K} \vect{\zeta})^\top (2 \vect{I} \vect{x} - \vect{K} \vect{\zeta} ) \nonumber\\& + \vect{\lambda}^\top (||\vect{y} - \vect{x}||_1 - \vect{r}). 
\end{align}
Finally, the dual problem can be written as:
\begin{align}
    \underset{\vect{\lambda}}{\text{maximize }} g(\vect{\lambda}) &= \underset{\vect{x}}{\text{inf}} \biggl\{ \nonumber\\ \nonumber &\vect{x}^\top (\gamma\vect{I}^\top \vect{I} + \alpha\vect{E}^\top \vect{E} + \beta \vect{R}^\top \vect{R}) \vect{x} \nonumber\\& - \gamma(\vect{K} \vect{\zeta})^\top (2 \vect{I} \vect{x} - \vect{K} \vect{\zeta} ) \nonumber\\& + \vect{\lambda}^\top (||\vect{y} - \vect{x}||_1 - \vect{r}) \biggr\} \\ 
    \text{subject to} \ &\vect{\lambda} \succeq 0, \; \alpha, \beta, \gamma \ge 0,  \nonumber
\end{align}
which can be solved to find the optimal dual values $\vect{\lambda}^*$. The use of these dual values is explained in Sec.~\ref{subsec:PA}.

    
\subsection{Interpretations}
The $\vect{E}$ and $\vect{R}$ matrices used here have multiple interpretations, some of which include: Edges and ribs in an elastic map, finite derivatives of functions, Tangent and Laplacian coordinates, smoothing regularization in optimization, and interpolation between points. In this work, they represent the edges and ribs of the elastic map, which are ``spring'' connections between consecutive points. By increasing or decreasing the stretching and bending stiffness of these springs, a different tuning of the elastic map can be found. Another interpretation of these matrices involves the first and second finite derivatives of a function. Alternatively, \cite{Ravichandar2019MCCB} uses these matrices to transform demonstrations from Cartesian coordinates to Tangent and Laplacian coordinates. These different coordinate systems represent different properties of the Cartesian demonstrations. The Tangent coordinates represent the tangent vectors along the demonstration, and Laplacian coordinates represent the curvature of the demonstration. In optimization, these terms can be used for smoothing regularization~\cite{boyd2004convex}. To penalize variation in a solution, one or more regularization terms are added to the objective function. These regularization terms penalize the difference from a solution term to its neighboring solution terms, promoting a more uniform and smoother solution. Using the matrix we have denoted as $\vect{R}$ can also be used for interpolation of noise-free data~\cite{murphy2012machine}. Given some data points, the unknown points between them can be assumed as an average of the neighbors in addition to some noise. Composing this assumption for all points results in the second-order finite difference matrix.

\subsection{Perturbation Analysis}
\label{subsec:PA}

The optimization problem in \eqref{eq:opt}, and in general any optimization problem written in the standard form (considered the \textit{primal}), can be used to form a Lagrangian \textit{dual} function. One of the main advantages of the dual problem is that it is concave, even when the primal is not convex. Another important property of this formulation is that the optimal value of the Lagrangian dual problem, denoted as $d^*$, is the best lower bound on the primal optimal value, denoted $p^*$. In our formulation, the optimal duality gap (the distance between $d^*$ and $p^*$) is zero, so we say that \textit{strong duality} holds. This is correct because our primal problem is convex.

For each constraint in an optimization problem, there is an associated optimal dual value related to the dual problem, with the optimal dual solution $\lambda_i^*$ relating to the $i$th constraint. These dual solutions are related to how the optimal value, denoted $p^*$, changes when that constraint is perturbed. Perturbing a constraint changes the constraint by some small value $u$. For example, the constraint $x_1 - x_2 \leq 0$ would be perturbed to $x_1 - x_2 \leq u$. The optimal value for the original and perturbed constraint is denoted as $p^*(0)$ and  $p^*(u)$, respectively. The following rules apply when perturbing constraints~\cite{boyd2004convex}:
\begin{itemize}
    \item If $\lambda_i^*$ is large and the $i$th constraint is tightened ($u < 0$) then the optimal value $p^*(u)$ increases greatly.
    \item If $\lambda_i^*$ is small and the $i$th constraint is loosened ($u > 0$) then the optimal value $p^*(u)$ will not decrease greatly.
    \item If $\lambda_i^*$ is 0 then the $i$th constraint has no effect on the optimal value.
\end{itemize}
For all problems with strong duality when Slater's condition is satisfied, the following global inequality holds for all $u$~\cite{boyd2004convex}.
\begin{equation}
    p^*(u) \geq p^*(0) - \lambda^*u. \label{eq:sensitivity}
\end{equation}
If the optimal value changes slightly, then the optimal solution also changes slightly. This can be leveraged when reproducing demonstrations to measure confidence in the reproduction in the presence of perturbations. Additionally, if the optimal value does not change at all, then the optimal solution does not change either. Therefore any constraints applied with $\lambda_i^* = 0$ are unnecessary and can be removed. 

\subsection{Constraints Formulation}
\begin{figure*}[ht]
\centering
    \includegraphics[width=0.32\linewidth]{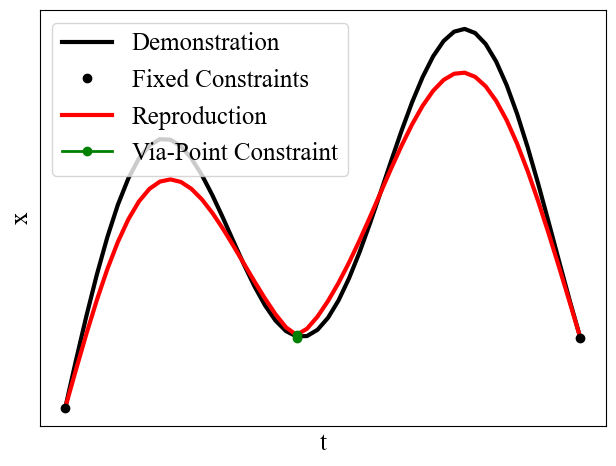}
    \includegraphics[width=0.32\linewidth]{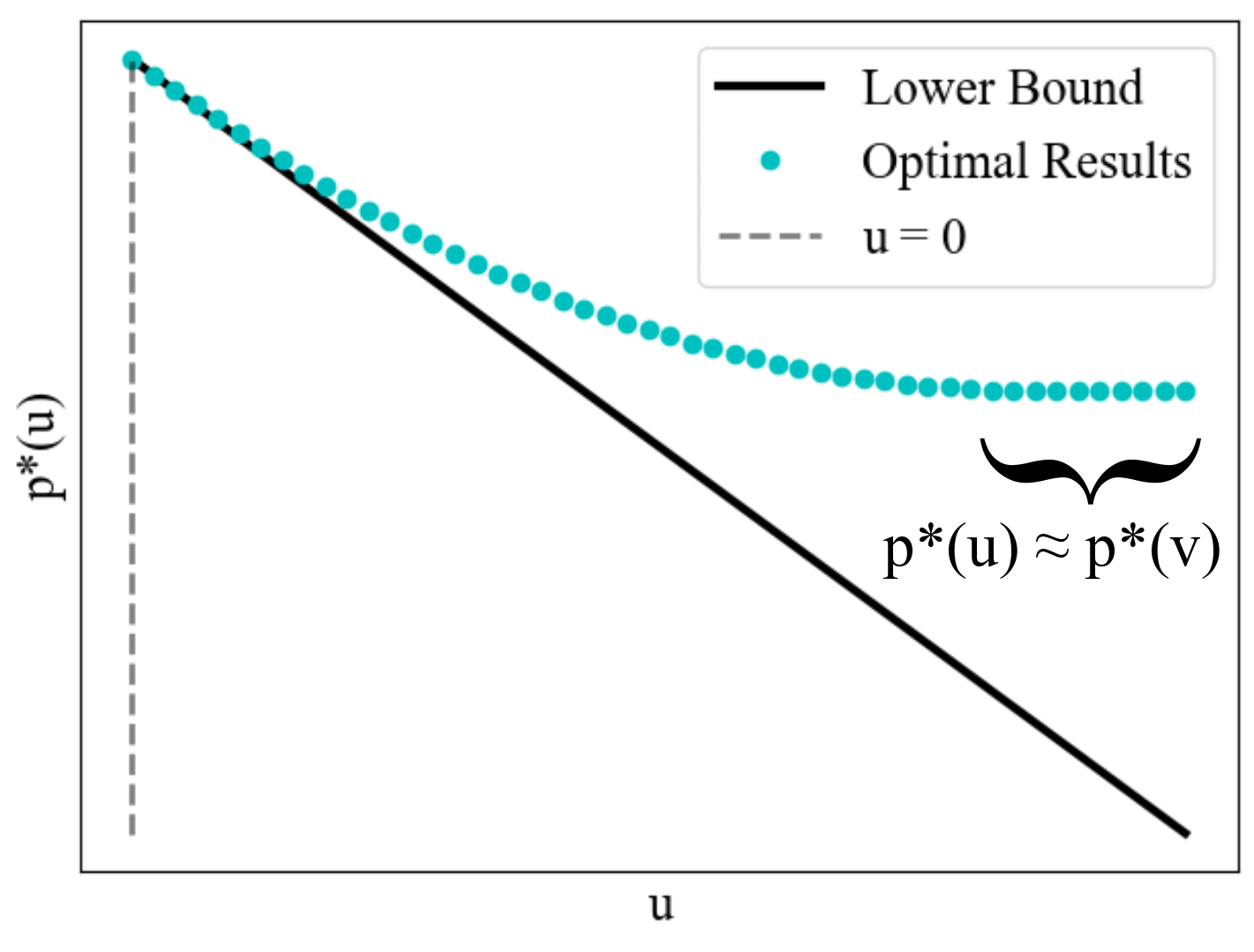}
    \includegraphics[width=0.32\linewidth]{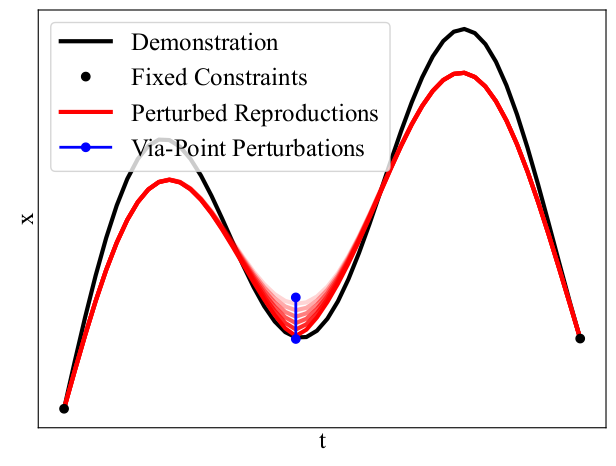}
    \caption{\small The perturbation analysis process shown in a simulated environment over a via-point constraint, with several reproductions of varying confidence calculated. Left: solution to the original problem with endpoint and via-point constraints. Center: perturbation analysis of the via-point constraint. The optimal value decreases when the constraint is loosened, leading to a smoother but less confident reproduction. Right: several reproductions of varying confidence levels. Confidence is shown with opacity.}
\label{fig:via_point_exp}
\end{figure*}

We consider constraints for initial, final, or via-points and/or obstacle avoidance. These are formulated as inequalities, which allows for flexibility in reproductions up to a given bound. 
We define a point constraint as
\begin{equation}
    f_i(x) = ||y - x_j||_1 - r,
\end{equation}
where $y$ is the point to constrain to, and $r$ is some safe radius around that point (see Fig.~\ref{fig:via_point_exp}). This forces a node $x_j$ to be within some radius $r$ around a specified point $y$. For $r=0$ this can be made an exact constraint. This formulation can work for initial, final, or via-point constraints. The perturbed version of this constraint is $f_i(x) \leq u$, 
where $u \geq -r$. This perturbs the safety radius around the constrained point, either tightening ($u \leq 0$) or loosening ($u \geq 0$) the constraint.

For obstacle avoidance constraints, we assume that the demonstrations have successfully avoided the obstacle. If this is not the case, new demonstrations can generated using methods such as TLFSD~\cite{hertel2021TLFSD}. Under this assumption, we create safety regions around the original demonstrations which avoid the obstacle. This is formulated as a point constraint, where $y = \zeta_j$, the closest point on the demonstration which has avoided the obstacle, and $r = ||\zeta_j - b||_1$, a safety radius from the demonstration to the obstacle, where $b$ is the closest obstacle. 
The full constraint is written as 
\begin{equation}
    f_i(x) = ||\zeta_j - x_j||_1 - ||\zeta_j - b||_1,
\end{equation}
This constraint is perturbed in the same way as point constraints, with the bounds $-||\zeta_j - b||_1 \leq u \leq 0$.

Additionally, we can use the property that if constraint $f_i(\vect{x})$ has $\lambda_i^* = 0$, it does not affect the optimal value or solution. Therefore, for obstacle avoidance constraints, we apply an obstacle avoidance constraint to each point along the reproduction, solve equation~\eqref{eq:opt}, and then remove all constraints with  $\lambda_i^* = 0$. This leaves only the constraints which affect the reproduction and change the reproduction when perturbed. Note that this method works for convex and non-convex obstacles.

\subsection{Confidence Measurement}

We wish to leverage the properties of perturbing constraints to measure confidence in reproductions. Given a reproduction generated under certain constraints, loosening one of those constraints would increase the confidence in that reproduction in the new environment, and vice versa for tightening constraints. To establish a scale for confidence, the tightest a perturbation can be is the lower bound, which is $-r$. As an upper bound for point constraints, there exists a region for which loosening the constraint no longer significantly decreases the optimal value, an example of which is shown in Fig.~\ref{fig:via_point_exp}. That is, $p^*(u) \approx p^*(v)$ for $v > u$. We denote the start of this region $u_{upper}$, and establish a scale for confidence $\sigma_c = (u + r)/(u_{upper}+r)$, such that $0 \leq \sigma_c \leq 1$. For obstacle constraints, we consider the upper bound the original $u=0$, as this is the loosest constraint which guarantees obstacle avoidance. We can measure the confidence using this scale of perturbations and given how $p^*$ changes with $\lambda_i^*$.

For obstacle avoidance constraints, there may be multiple constraints which need to be perturbed (i.e., in the case where two or more different points along a reproduction are in near collision with an obstacle. In this case, we apply a confidence factor $\sigma_s$ to all obstacle avoidance constraints, where $0 \leq \sigma_s \leq 1$. A higher confidence factor tightens the constraints and increases confidence in reproductions, whereas a lower confidence factor loosens the constraints. Using this factor, $u$ values are generated such that $u = -\sigma_s||\zeta_j - b||_1$. This confidence factor is selected according to the user.

\section{Experiments}
\label{sec:exps}

\subsection{Experimental Setup}

We validate our approach using several simulated and real-world experiments. We analyze different constraints, the perturbation of constraints, and how these perturbations relate to confidence in reproductions. All experiments are performed using code\footnote{ \url{https://github.com/brenhertel/LfD-Perturbations}} written in Python 3.8 utilizing the cvxpy library~\cite{diamond2016cvxpy}. Uniform weights (i.e., $w_1 = w_2 = ... = w_n$) are used for the approximation energy, and the bending and stretching constants are tuned manually.

\subsection{Perturbing a Via-Point Constraint}

We first validate our approach in a simulated reaching environment. In this experiment, shown in Fig.~\ref{fig:via_point_exp} (left), a demonstration is given that reaches from an initial point towards a specified endpoint, but a via-point is given with tight upper and lower limits. In this experiment, we only perturb around a single axis as shown in Fig.~\ref{fig:via_point_exp} (right), but multiple axes can be used for perturbations. Additionally, the bounded region is tight but non-zero.Using $r=0$ generates values for $\lambda_i^*$ which are inaccurate. 
A reproduction is generated using this set of constraints, shown in red in Fig.~\ref{fig:via_point_exp} (left). We wish to analyze how the reproduction may change if the constraint is perturbed. Multiple $u$ values are generated which loosen the constraint. The result of these perturbations on the optimal value is shown with cyan dots in  Fig.~\ref{fig:via_point_exp} (center). Loosening the constraint decreases the cost of the optimal reproduction, resulting in smoother reproductions. These costs remain above the lower bound set by inequality in~\eqref{eq:sensitivity}, shown in black. The constraint is loosened until increasing $u$ would no longer decrease $p^*(u)$. These perturbed reproductions are plotted in  Fig.~\ref{fig:via_point_exp} (right), where opacity corresponds to confidence. Reproductions found under tighter constraints are considered more confident than those with looser constraints. The reproduction found under the tightest constraints is considered to have a confidence of 1, whereas a reproduction generated under the loosest constraints (when loosening the constraint no longer decreases the optimal value) has a confidence of 0. This experiment shows how confidence in reproductions can be measured by perturbing a constraint.

\begin{figure}[ht]
    \centering
    \includegraphics[width=0.49\columnwidth]{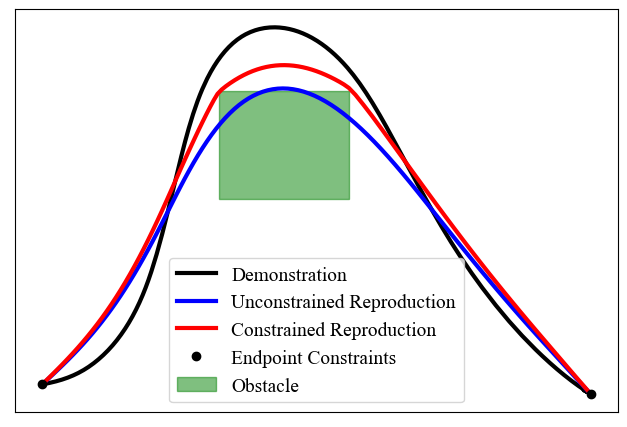}
    \includegraphics[width=0.49\columnwidth]{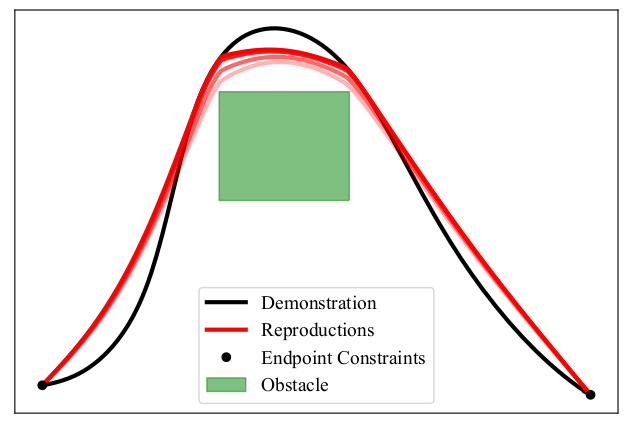}
    \caption{\small Left: optimal reproduction for an obstacle avoidance task with (red) and without (blue) constraints for obstacle avoidance. Right: reproductions of varying levels of confidence, where confidence is shown with opacity.}
    \label{fig:obs_exp}
\end{figure}

\subsection{Perturbing Obstacle Avoidance Constraints}

In this experiment, we show how constraints can be applied to obstacle avoidance and how confidence can be measured in obstacle avoidance skills. This experiment, shown on the left in Fig.~\ref{fig:obs_exp}, involves a demonstration (black) reaching around an obstacle (green). Simply generating a reproduction does not successfully avoid the obstacle (blue), so obstacle avoidance constraints are applied, resulting in a reproduction which narrowly avoids the obstacle (red). Several confidence factors ranging from 0 to 1 are shown on the right of Fig.~\ref{fig:obs_exp}, with the confidence factor shown in opacity. It can be seen that as the confidence factor increases, the reproduction avoids the obstacle by a larger margin. The confidence factor directly correlates to confidence in the reproduction, with higher factors more similar to the demonstration, but more constrained. Some reproductions of very high confidence contain jagged corners, high-jerk features which may be undesirable in robot reproductions.

\subsection{Perturbing Reproductions in the Real World}

\begin{figure}[ht]
    \centering
    \includegraphics[width=0.9\columnwidth]{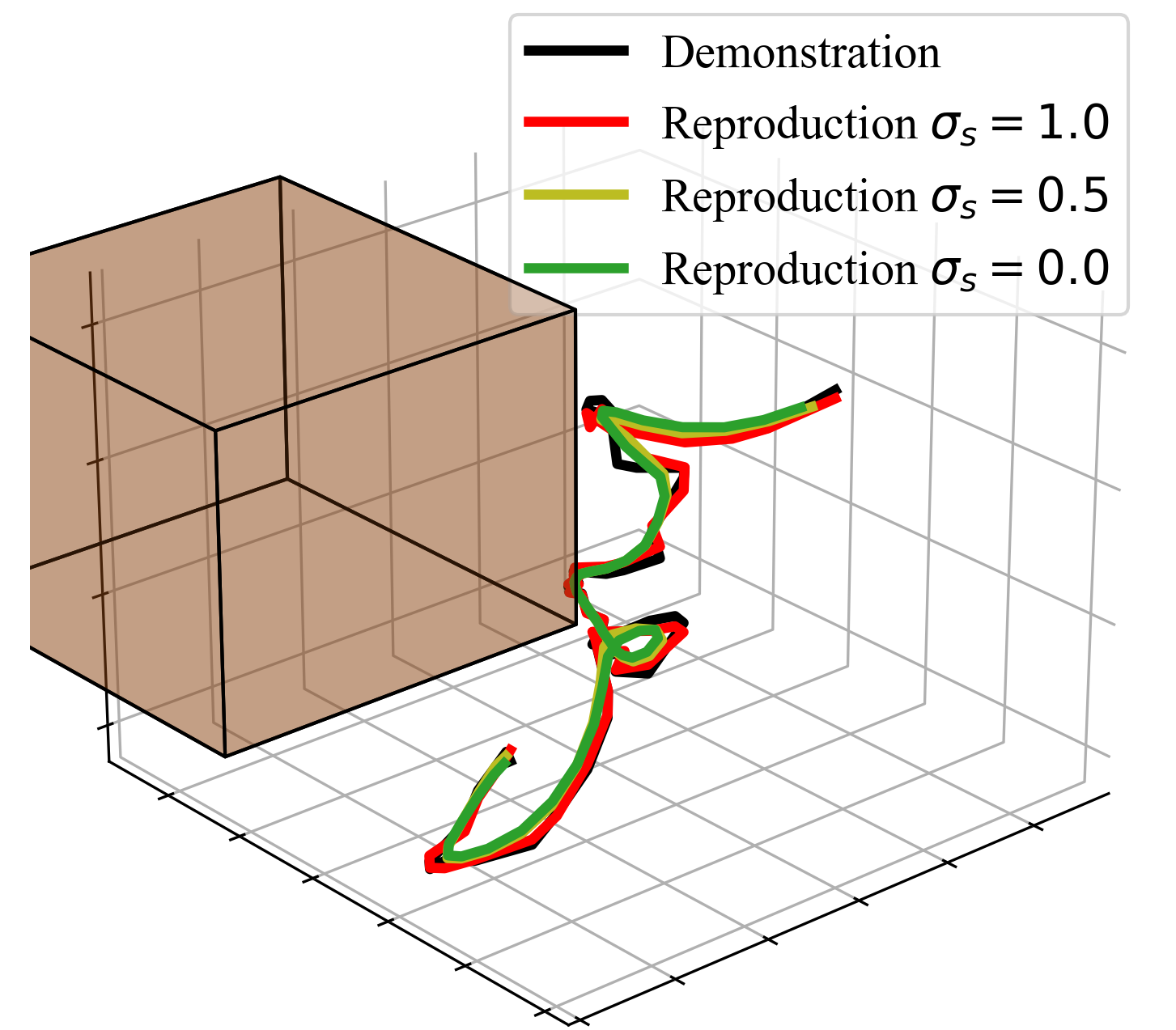}
    \includegraphics[width=0.45\columnwidth]{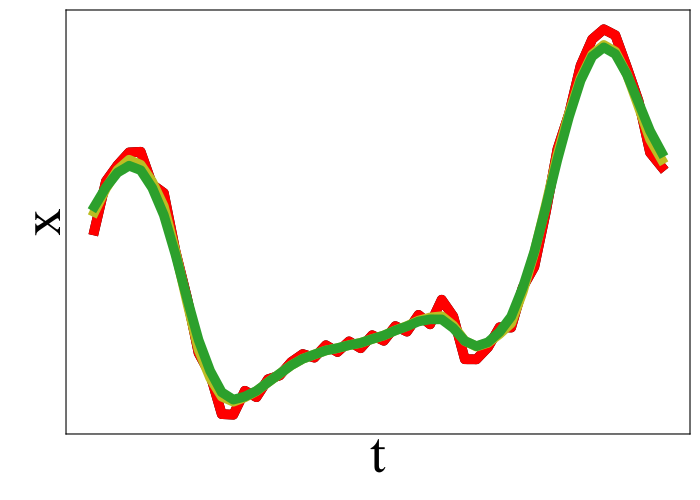}
    \includegraphics[width=0.45\columnwidth]{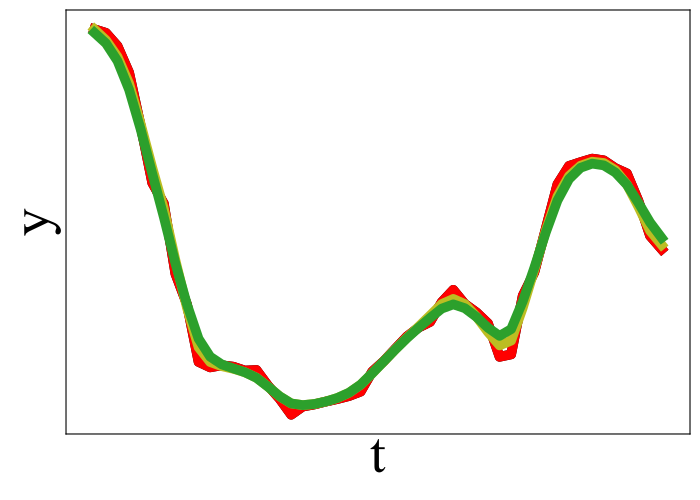}
    \caption{\small Demonstration and reproductions with different confidence factors of opening a real-world box. Individual x and y dimensions are shown to highlight differences in reproductions. Higher confidence factors generalize less, but are more confident in the ability for the reproduction to successfully complete the skill.}
    \label{fig:box_exp}
\end{figure}

Finally, we use a real-world Kinova Jaco2 7DOF manipulator arm in a door opening skill. An electrical box (shown in Figs.~\ref{fig:fig1}, \ref{fig:fig2}, and represented in brown in Fig.~\ref{fig:box_exp}) is placed in front of the arm which must be opened. A human demonstrates the task through kinesthetic teaching, where the arm is guided towards the edge of the box and pulled back to open the door.\footnote{See accompanying video: \url{https://youtu.be/IQDxbhEiNbk}} The demonstration successfully completes the skill, but is jagged and suboptimal. Therefore, we wish to establish a level of confidence which is able to successfully complete the skill while incorporating features of the LfD representation such as smoothness and generalization. We apply obstacle avoidance constraints and generate three reproductions with confidence factors 0, 0.5, and 1, shown in Fig.~\ref{fig:box_exp}. All reproductions are computed a priori and executed in the same real-world environment the demonstration was recorded, where the reproductions with confidence levels of 0.5 and 1 successfully reproduce the task. Of these successful reproductions, the lower confidence level is more desirable, as it is smoother than the reproduction with a confidence factor of 1. Higher confidence levels, while more successful at completing tasks, are tightly constrained which may cause undesirable features in reproductions, such as the high-jerk movements seen in the $\sigma_s = 1$ reproduction shown in Fig.~\ref{fig:box_exp}. This shows how different confidence levels can be used to find tradeoffs between desired features in the reproduction and demonstration while successfully completing the intended skill.

\section{Conclusions and Future Work}
\label{sec:conclusion}

We have presented a method for finding confidence in reproductions through perturbation analysis. We have shown how the estimated confidence values can be used to inform users or future algorithms about various properties of reproductions and the safety of trajectory execution. Additionally, We have validated the utility of this technique through several experiments, including using via-point generalization and obstacle avoidance constraints. Beyond what we have shown here, there are several opportunities for future work. Firstly, we measure confidence by comparing a reproduction to the demonstration, assuming the demonstration to be the ``most confident.'' This may not always be the case because demonstrations may be noisy or otherwise sub-optimal. Instead, information from the surrounding environment could be used to find a better measure of confidence. Another avenue for future work is online adaptation of trajectories based on confidence. This paper considers stationary environments with trajectories computed a priori and executed with a low-level non-reactive controller. Real-world environments, however, are non-stationary and challenging, and require robots to adapt on-the-fly to the dynamic changes. 

\section*{Acknowledgements}
This work was supported by the U.S. Office of Naval Research (N00014-21-1-2582). Additional thanks to Ryan Donald for his assistance.
\typeout{}
\bibliographystyle{IEEEtran}
\bibliography{references}

\end{document}